\documentclass[10pt,journal,compsoc]{IEEEtran}
\ifCLASSOPTIONcompsoc
  \usepackage[nocompress]{cite}
\else
  \usepackage{cite}
\fi
\ifCLASSINFOpdf
\else
\fi
\usepackage{url}
\usepackage{graphicx}
\usepackage{epstopdf}

\usepackage{hyperref}
\usepackage{footnote}
\usepackage{balance}
\usepackage{amssymb}

\usepackage{amsmath}
\usepackage{multicol}
\usepackage{multirow}
\usepackage{anyfontsize}
\usepackage{array}
\usepackage{hhline}
\usepackage{float}
\usepackage{enumitem}
\usepackage{tabularx}

\begin{document}
%
\title{Hyper360 - a Next Generation Toolset\\ for Immersive Media}
%
%
%
%

\author{Hannes~Fassold, Antonis~Karakottas, Dorothea~Tsatsou, Dimitrios~Zarpalas, Barnabas Takacs, Christian Fuhrhop, Angelo Manfredi, Nicolas Patz, Simona Tonoli, Iana Dulskaia

\IEEEcompsocitemizethanks{
\IEEEcompsocthanksitem H. Fassold is with JOANNEUM RESEARCH - DIGITAL, Austria.

\IEEEcompsocthanksitem A. Karakottas, D. Tsatsou, D. Zarpalas are with Centre for Research \& Technology Hellas, Greece.
\IEEEcompsocthanksitem B. Takacs is with Drukka Kft/PanoCAST, Hungary.
\IEEEcompsocthanksitem Christian Fuhrhop is with FRAUNHOFER FOKUS, Germany.
\IEEEcompsocthanksitem Angelo Manfredi is with Engineering Ingegneria Informatica S.p.A., Italy.
\IEEEcompsocthanksitem Nicolas Patz is with Rundfunk Berlin-Brandenburg, Germany.
\IEEEcompsocthanksitem Simona Tonoli is with R.T.I. S.p.A. – Mediaset S.p.A, Italy.
\IEEEcompsocthanksitem Iana Dulskaia is with Eurokleis s.r.l, Italy.

}
\thanks{Manuscript received April 19, 2005; revised August 26, 2015.}}

%
%

\markboth{Journal of \LaTeX\ Class Files,~Vol.~14, No.~8, August~2015}%
{Shell \MakeLowercase{\textit{et al.}}: Bare Demo of IEEEtran.cls for Computer Society Journals}
%



\IEEEtitleabstractindextext{%
\begin{abstract}
Spherical 360$^\circ$ video is a novel media format, rapidly becoming adopted in media production and consumption of immersive media. Due to its novelty, there is a lack of tools for producing highly engaging interactive 360$^\circ$ video for consumption on a multitude of platforms. In this work, we describe the work done so far in the Hyper360 project on tools for mixed 360$^\circ$ video and 3D content. Furthermore, the first pilots which have been produced with the Hyper360 tools and results of the audience assessment of the produced pilots are presented.
\end{abstract}

}

\maketitle

\IEEEdisplaynontitleabstractindextext

%
\IEEEpeerreviewmaketitle

\IEEEraisesectionheading{\section{Introduction}\label{sec:introduction}}

%
%
%
%
\IEEEPARstart{E}{xtended} Reality (XR), which comprises Virtual Reality (VR), Augmented Reality (AR) and Mixed reality (MR), creates entirely new ways for consumers to experience the world around them and interact with it. Within the last few years, improvements in sensor technology and processing power have led to tremendous advances in all aspects of XR hardware, and due to economies of scale of the massively growing XR market, these devices are available now at a reasonable price point. On the production side, powerful low-cost systems for capturing 3D objects, volumetric video and 360$^\circ$ videos make it possible to create budget VR/AR productions. The same applies to the consumption side, where VR headsets like the Oculus Go or Playstation VR provide a highly immersive VR experience, which is affordable for everyone.

Unfortunately, the development of tools and technologies for authoring, processing and delivering interactive XR experiences is lagging considerably behind the hardware development, which is a hurdle for the cost-effective mass production of appealing XR content and scenarios. Lack of content, in turn, hinders broader adoption and acceptance of XR technologies by the consumer. For all these aspects, new technologies and tools are needed in order to overcome the specific challenges (like multimodal data, non-linear interactive storytelling, annotation and metadata models) of XR content creation.

The Hyper360 project (\emph{\url{https://www.hyper360.eu}}) aims to fill this gap by providing a convenient toolset for XR media, with a focus on mixed 360$^\circ$ video and 3D content. We address the whole workflow, covering capturing, production, delivery and content consumption on a variety of devices. Through augmentation of the 360$^\circ$ video with 3D content, novel opportunities for storytelling are provided. Additionally, we provide a personalized consumption experience via extracting and adapting to the viewer preferences. The user partners in the project have produced pilots with the tools and performed audience assessment sessions with those pilots.

In the following, we give an overview of the tools which have been developed in the Hyper360 project, grouped according to their respective workflow phases (from capturing to consumption). Furthermore, the produced pilots and key results of the audience assessment are presented. 

\section{Capturing Tools}
\label{sec:capturing}

For capturing, we developed the \emph{OmniCap} tool, which is responsible for capturing 360$^\circ$ video with a variety of cameras. 
In order to ensure high quality content, a novel quality analysis component has been integrated into the OmniCap tool. Furthermore, several AI components have been added for extracting semantic information from the video.

\begin{figure}[t]
	\centering
		\includegraphics[width=0.4\textwidth]{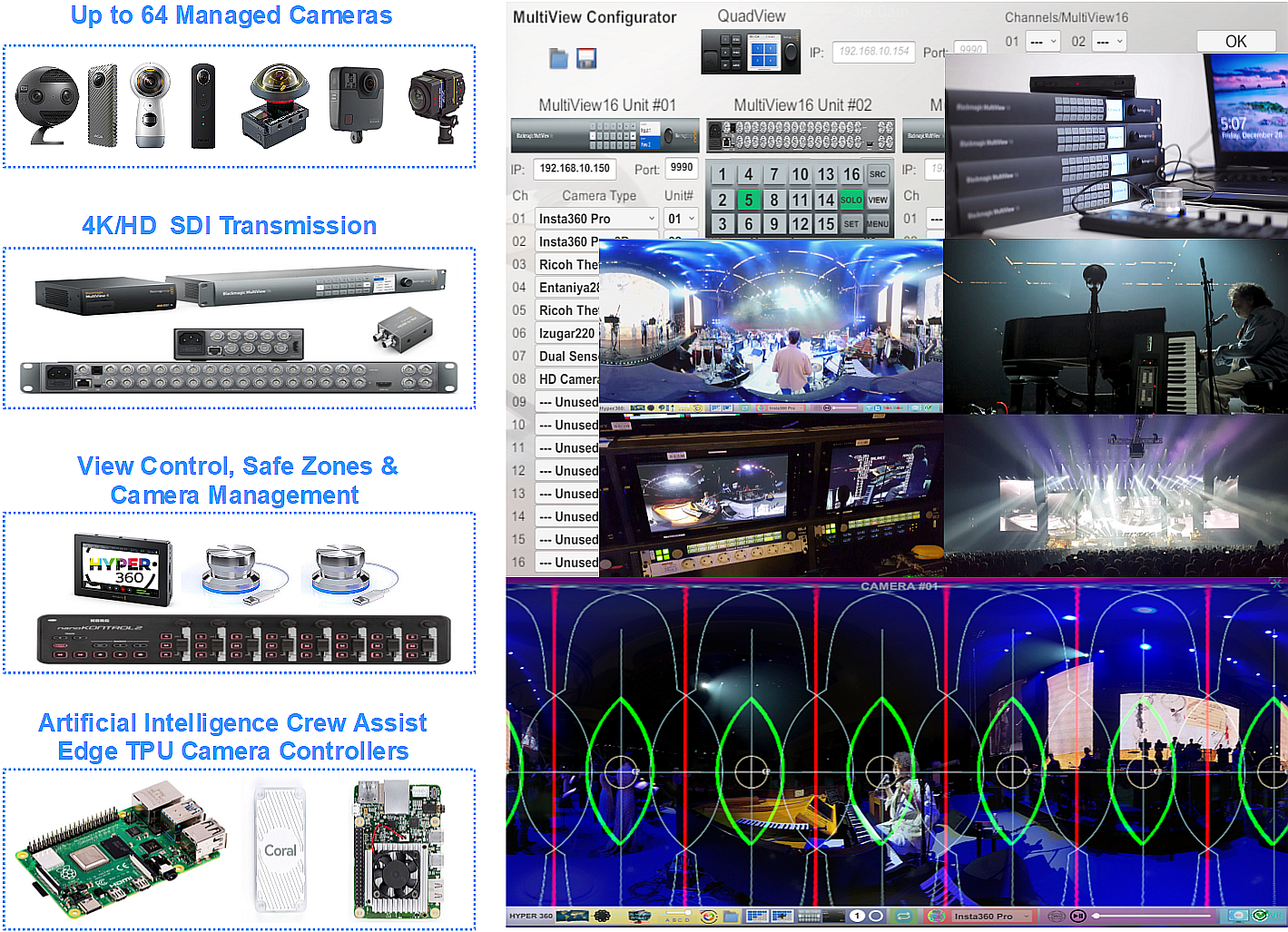}
	\caption{OmniConnect capturing tool} 
	\label{fig:omniconnect}
\end{figure}

OmniCap supports a number of modern 360$^\circ$ camera rigs (including multi camera arrays or tiny fisheye lenses devices) along with regular HD cameras in a generic architecture that can support the remote management of up to 64 mixed camera units and configurations. It was designed with both multi-camera 360$^\circ$ as well as 3D captures to support Free Viewpoint Video suited for the broadcast sector. Technically, our solution is based on a generalized inverse fisheye lens transformation implemented in real-time, whereas each individual camera view is first distorted in real-time and subsequently combined into a single equirectangular view that can be viewed in flat mode (seeing the entire scene in a distorted manner as shown in Figure \ref{fig:omniconnect}) or mapped onto a virtual sphere to be able to frame and look at elements within the scene. Each camera’s input is mapped onto a graphic element (patch) and multiple patches are blended and distorted to create seamless transitions in the overlapping areas. Since the purpose of this tool is on-set pre-visualization, control and quality check to find the best camera setup, it also includes additional overlay elements such as the visualization of safe-zones.
For quality control, existing algorithms for detection of defects like signal clipping, blurriness, flicker or noise level on conventional video have been adapted to the specifics of 360$^\circ$ video (especially the equirectangular projection) and extended in order to provide localized defect information. General
strategies how to adapt computer vision algorithms to the specifics of 360$^\circ$ video (like the equirectangular projection) are given in \cite{Fassold2019Adapting}. Furthermore, OmniCap integrates a number of AI modules in order to provide rich semantic high-level information to other tools in the Hyper360 toolset. Specifically, it integrates modules for automatic detection and tracking of objects, instance segmentation, facial landmark extraction, hair mask segmentation and monoscopic depth estimation.

\section{Post-processing Tools}
\label{sec:production}

For post-processing, \emph{OmniConnect} has been developed for annotation / enrichment of 360 $^\circ$ media. The \emph{CapTion} tool handles 3D human performance capturing and embedding in the 360$^\circ$ video.

\begin{figure}[t]
    \centering
    \includegraphics[width = 0.45\textwidth]{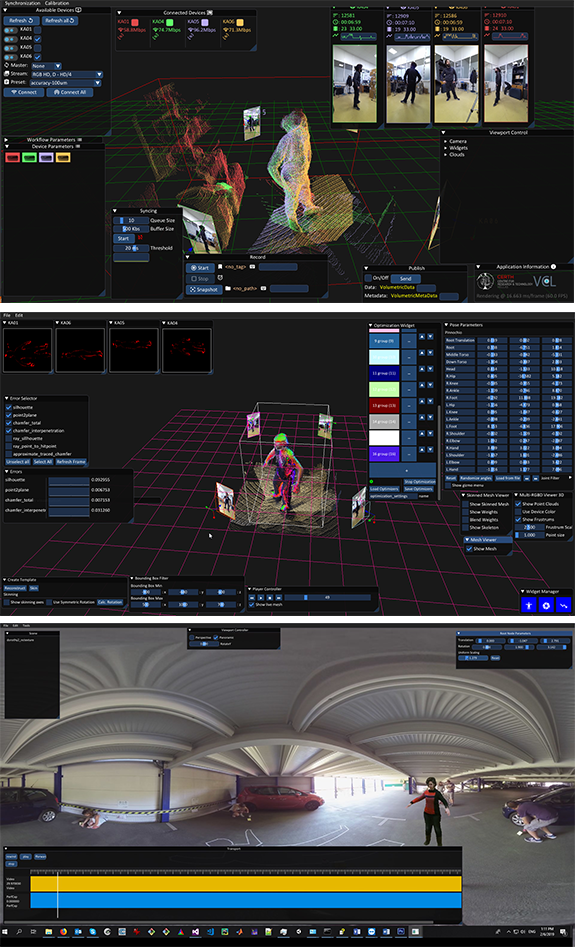}
    \caption{Components of the CapTion tool for volumetric capture, 3D production and fusion}
    \label{fig:Caption}
\end{figure}

The \emph{OmniConnect} tool is a browser-based authoring tool for the enrichment of 360 $^\circ$ video with hotspots, for which certain events can be defined. It supports different types of hotspots like shapes, 2D video, audio, HTML pages, multline text and images. For each hotspot, its properties (like position, color, view option) can be changed easily in a graphical interface. A player is provided inside OmniConnect for previewing the enriched video. Internally, it employs efficient 3D frameworks like three.js and WebGL for real-time rendering of the enriched 360$^\circ$ video in the player. 

\begin{figure}[t]
    \centering
    \includegraphics[width = 0.45\textwidth]{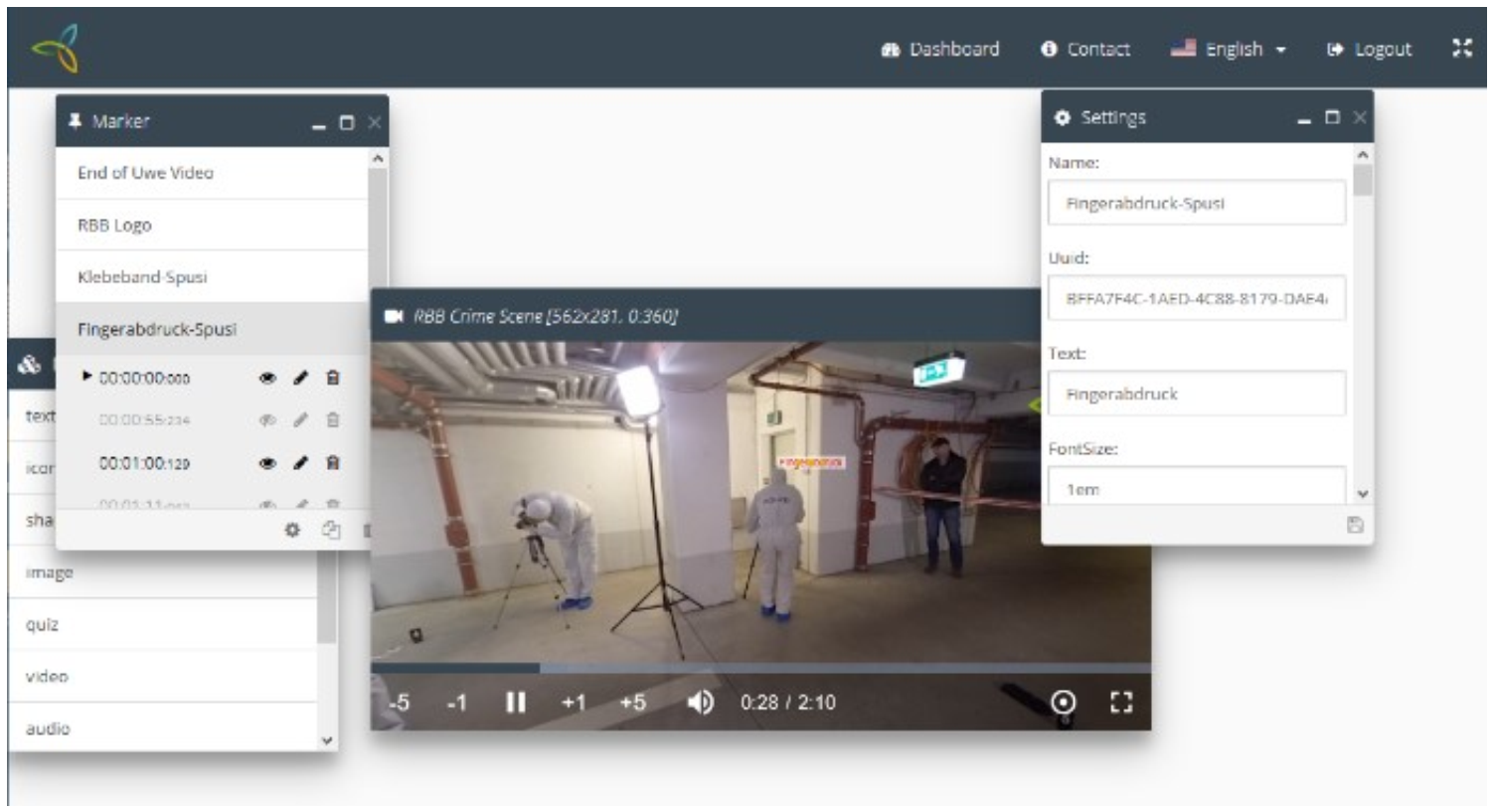}
    \caption{Editing tool for HbbTV.}
    \label{fig:hbbtv_editor}
\end{figure}

Due to the different requirements of authoring for interactive TV (HbbTV) compared to smartphones and HMDs, Hyper360 uses a different editor for interactive TV applications. Visual elements, such as icons, pictures and videos are re-purposed for this editor, but the user interaction with the 360$^\circ$ video and interactive elements will be authored according to the specific requirements of TV usage. Figure \ref{fig:hbbtv_editor} illustrates the graphical interface for the editor for HbbTV.

Volumetric video capturing of human performances digitizes
real performers and allows for their compositing into in 360$^\circ$
media. The \emph{CapTion} tool covers all aspects of such productions by
providing an end-to-end pipeline for 3D capturing, animated
3D asset creation and 3D-360$^\circ$ fusion. It comprises a portable and affordable volumetric capturing component, using the latest consumer level RGB-D sensor technology, a 3D production component based on an advanced human
digitization technology that converts recorded multiview RGB-D sequences of human performances to animated 3D assets and a mixed media (3D and 360$^\circ$) fusion component that allows for the realistic compositing of animated 3D objects into
360$^\circ$ videos. The \emph{capturing component} is a multi-view system developed in a distributed client-server architecture, where server software is deployed on acquisition nodes with each node being composed of an RGB-D sensor and a miniPC, while the client is deployed on a main workstation and is responsible for orchestrating the process of capturing and recording. We utilize the latest advances in RGB-D sensor technology employing both the Intel Realsense D400 sensors, as well as the recently released Microsoft Azure Kinect. It provides
 the state-of-the-art method of \cite{sterzentsenko2020deepsoftprocrustes} for volumetric sensor alignment (i.e. extrinsic sensor calibration), that leverages low-cost packaging boxes and allows arbitrary camera placement configurations. The \emph{3D production component} transforms the volumetrically aligned RGB-D streams into an animated 3D asset. It employs automated performance capture technology, which requires no external photogrammetry or rigging software, which is typical for other systems. Apart from lowering the cost and technical barrier, this also smooths the workflow. The template creation part that involves creating the actor’s 3D model and associated metadata required for animating it (bone hierarchy and skinning weights), albeit optional, is integrated and automated into the tool. It is also possible to reuse templates among captures, or even use an externally provided one. The \emph{fusion component} leverages state-of-the art AI algorithms for depth \cite{zioulis2019spherical}  and surface normal estimation \cite{karakottas2019360}, in order to provide a seamless blending of the 3D content with the 360$^\circ$ video. In Figure \ref{fig:Caption} the individual components are shown.
 
\section{Delivery Tools}
\label{sec:delivery}

The delivery tools comprise cloud services that facilitate the delivery of produced, integrated media as seamless, personalised 360$^\circ$ experiences. Specifically, it includes user profiling and recommendation services for personalized consumption and \emph{automatic cinematography} for generating a 2D video for playout on a conventional TV set. 

\begin{figure}[t]
    \centering
    \includegraphics[width = 0.45\textwidth]{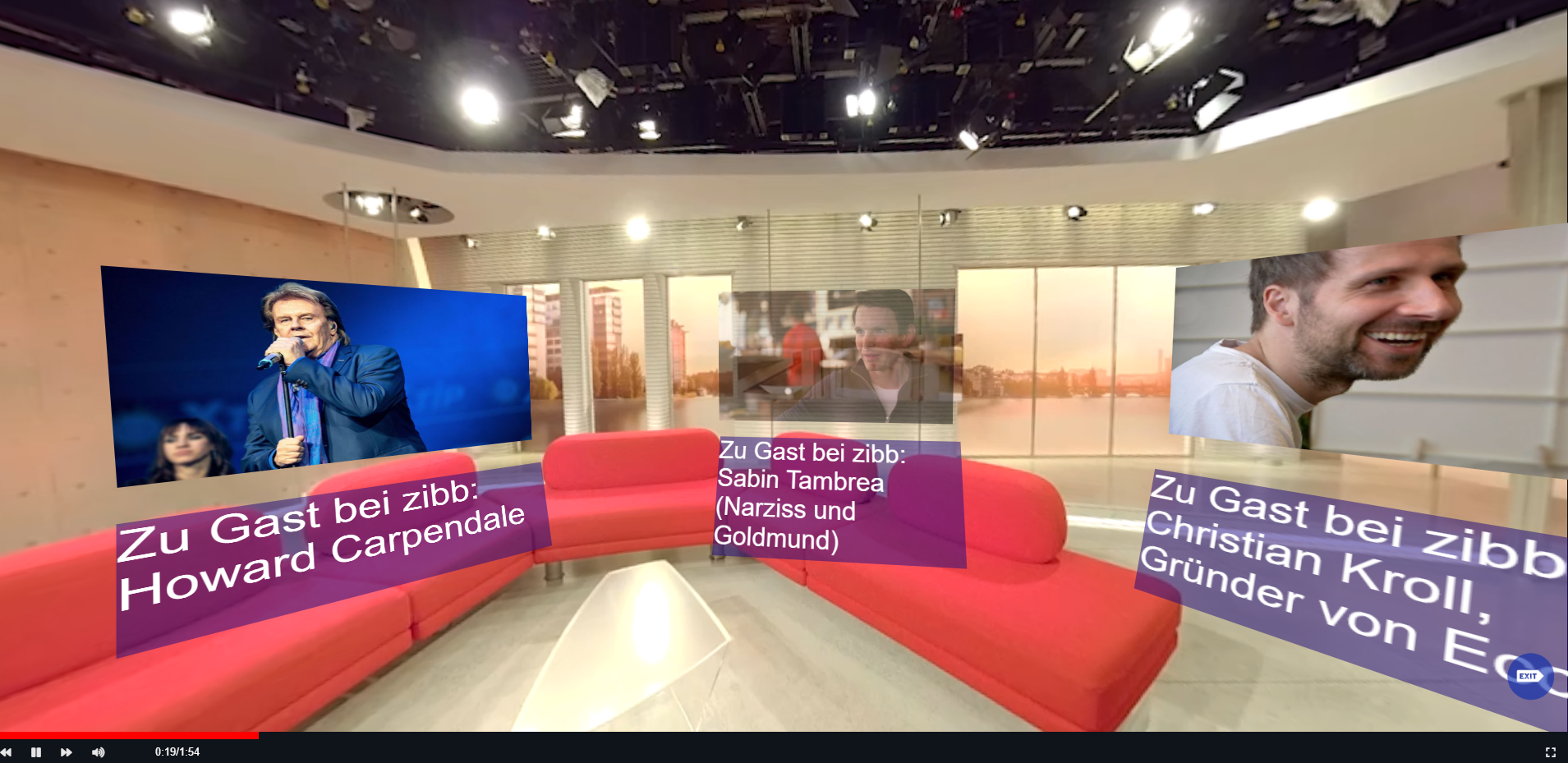}
    \caption{. Example of embedded objects recommendation in omnidirectional video. Opacity designates the level of preference the user has in each object.}
    \label{fig:Recom}
\end{figure}

The \emph{personalisation mechanism} relies on the holistic representation of content annotations, in order to minimise loss of information and characterise it with enhanced metadata. To this end, through Hyper360's \emph{Semantic Interpreter}, content annotation is construed to a set of fuzzy concept instances, based on an expanded version of the LUMO ontology \cite{tsatsou2014lumo}. This interpretation builds upon learned lexico-syntactic relations between free-form, domain-specific metadata. Subsequently, the \emph{Profiling Engine} harvests the novel opportunities that omnidirectional video offers for implicitly capturing the viewers’ preferences. The viewpoint choice allows the \emph{Profiling Engine} to capture spatio-temporal and other behavioural information (which part of the video the viewer focuses on and for how long, which objects in a scene they interact with, etc.), which are combined with appropriate semantic meaning of the content. Lastly, the \emph{Recommendation Engine} semantically matches learned user profiles with semantic content metadata, through an extension of the LiFR fuzzy reasoner \cite{tsatsou2014lifr}. The application of recommendation is three-fold within Hyper360, offering personalised navigation within the 360$^\circ$ media, which is manifested both as cues for personalised camera paths, as well as cues for personalised 3D Mentor narratives, while also achieving targeted embedded objects (hotspots, hyperlinks, nested media) delivery, as seen in Figure \ref{fig:Recom}.

The goal of \emph{automatic cinematography} is to calculate automatically a visually interesting camera path from a 360$^\circ$ video, in order to provide a traditional TV-like consumption experience. This is necessary for consumption of a 360$^\circ$ video on older TV sets, which do not provide any kind of interactive players for 360$^\circ$ video. Furthermore, even on devices capable of consuming 360$^\circ$ videos interactively, an user might prefer a lean-back mode, without the need to navigate actively to explore the content. The algorithm works in an iterative fashion, shot by shot. The shot length is set by the user, with a default of three seconds. Within the shot range, all objects (e.g. human, dog, cat, bicycle, car) appearing in the scene are detected and tracked throughout the shot with the deep learning based method proposed in \cite{Fassold2019Omnitrack}. For each scene object, a set of measures is calculated, like the average size, average motion magnitude, \emph{neighbourhood score} (indicating how isolated the object is)  and \emph{visited
score} (indicating how visible the object was in the most recent shots). In the next phase, a set of \emph{shot hypotheses} is calculated. In order to match the diversity of traditional film which employs different shot types (from close up to very wide shots) for artistic purposes, we employ also several shot types. Specifically, we support the following shot types: tracking shot, static shot, medium shot, pan shot and recommender shot. Each shot type is aimed at a certain purpose, differing in how the saliency scores for the objects are calculated, and has a certain style in terms of artistic elements like FOV (field of view) and camera movement. For example, the tracking shot is focused on objects which are moving, large and isolated (like the singer of a band) and tracks them throughout the shot. On the other hand, for a static shot the camera is static and the focus is on a group of objects, like the audience in a concert. They differ also in their FOV. For the tracking shot we employ a standard FOV of 75$^\circ$, whereas for a static shot a wider FOV of 115$^\circ$ is used in order to capture groups better. For each of these shot types, a couple of shot hypotheses (typically 2 – 4) are generated. For each shot hypothesis, a corresponding score is calculated which indicates how “good” this shot hypothesis is. The shot hypothesis score is influenced by a number of factors, like the saliency scores of the scene objects and rules of continuity editing (like the jump-cut rule for avoiding jump-cuts). The saliency score of an object indicates how "interesting" the object is and is dependent on several factors like shot type, object class, object size, motion magnitude, neighborhood score and visited score. All generated shot hypotheses are collected in a set, and the shot hypothesis with the highest score is now
chosen as the best one. All frames of the current shot are now rendered, based on the viewport information provided in the best shot hypothesis. In order to not let a certain shot type dominate the generated video, we implemented a mechanism which limits the occurrence of each shot type in the generated video. 

\begin{figure}[t]
	\centering
		\includegraphics[width=0.45\textwidth]{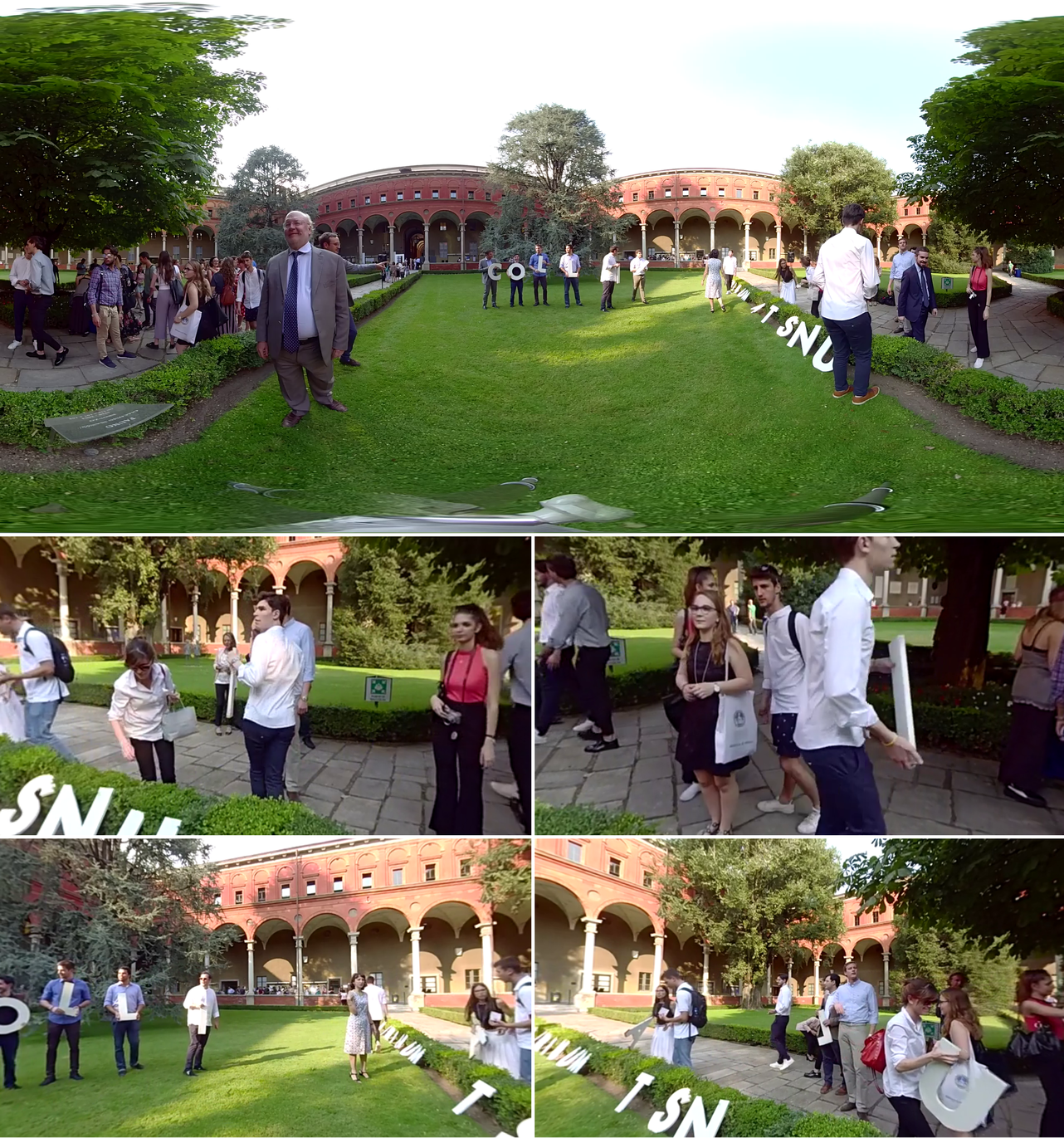}
	\caption{Result of automatic cinematography. First row shows input video, second and third show the first two generated shots.} 
	\label{fig:autocam_result}
\end{figure}

\section{Consumption Tools}
\label{sec:consumption}

%
%

\begin{figure}[t]
	\centering
		\includegraphics[width=0.45\textwidth]{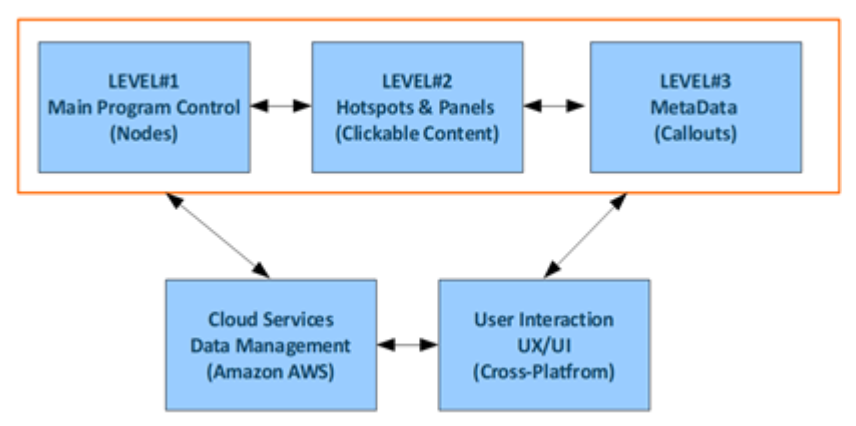}
	\caption{Control architecture and levels for OmniPlay unified player solution.} 
	\label{fig:player_control_files}
\end{figure}

The \emph{OmniPlay} unified player solution is a suite of cross-platform applications implemented on different platforms while adhering to the specific characteristics and requirements of those platforms. Each of these apps is based on a set of services offered by the underlying operating system, where both HW performance, SW limitations on video playback and device-user interaction
possibilities need to be considered separately for each case. While IOS and HbbTV clearly offer different viewer experiences, the Hyper360 player suite follows a unified control architecture to remove these implementation details at the authoring tool level and result in publishable control files affecting the flow and interaction patterns as depicted in Figure \ref{fig:player_control_files}.
In the following, we describe briefly the implementation and particular characteristics for each of the currently available platforms, namely mobile IOS, Android, Desktop / Web, VR Headsets and HbbTV. The \emph{Android} version of the Hype360 player is taking advantage of Unity3D cross compilation possibilities. In order to make the player mobile VR-ready, we integrated Google's VR component and services controlling head movement via the gyro sensors of the device. This allows the player to accommodate various screen sizes and low-cost googles into which viewers can place their smartphone and readily explore 360 experiences. The \emph{IOS} player employs the same codebase as the Android player, but was developed with XCode. The implementation for the \emph{Windows Desktop} player is based on the Unity3D 360$^\circ$ video playback code extracted from the OmniCap tool and extended with specific interactive features and XML program/track/callout management. For the \emph{web} player, the code for the preview functionality of the OmnniConnect tool was used as a basis. The player for \emph{VR headsets} (supporting HTC Vive and Occulus Go) is based on the Windows Desktop player modified with specific interaction modules to handle the peripheries and interfaces of the particular device. In particular, this effected software modules involved in virtual camera rotation, pointing device for selecting menu elements and hotspots with a 'teleport' interface, and of course rendering separate images for both eyes together with the locking/unlocking of views for flat video playback. Because the HTC Vive tracks 6DOF User motion, we also needed to attach
the video sphere directly to the head (parenting) and eliminate translational effects. For \emph{HbbTV}, a cloud-based approach enables the playback of 360$^\circ$ videos on a wide range of client devices - especially HbbTV-enabled terminals for seamless integration with broadcast and broadband TV services. It also addresses low performance devices like set-top-boxes and streaming devices like FireTV and Chromecast that do not have the capability to process 360$^\circ$ videos locally.

\section{Pilots and Audience Assessment}

The user partners in the Hyper360 project, the broadcasters Rundfunk Berlin-Brandenburg (RBB) and Mediaset (RTI), employed the first prototypes of the developed tools in order to produce pilot content. For this, RBB focused on a \emph{Immersive Journalism} scenario, which allows first person experience of the events or situations described in news reports and documentary film. 
The focus of RTI was on a \emph{Targeted Advertising} scenario, where entertainment or infotainment content is enriched with advertising objects and hyperlinks. 

\begin{figure}[t]
	\centering
		\includegraphics[width=0.45\textwidth]{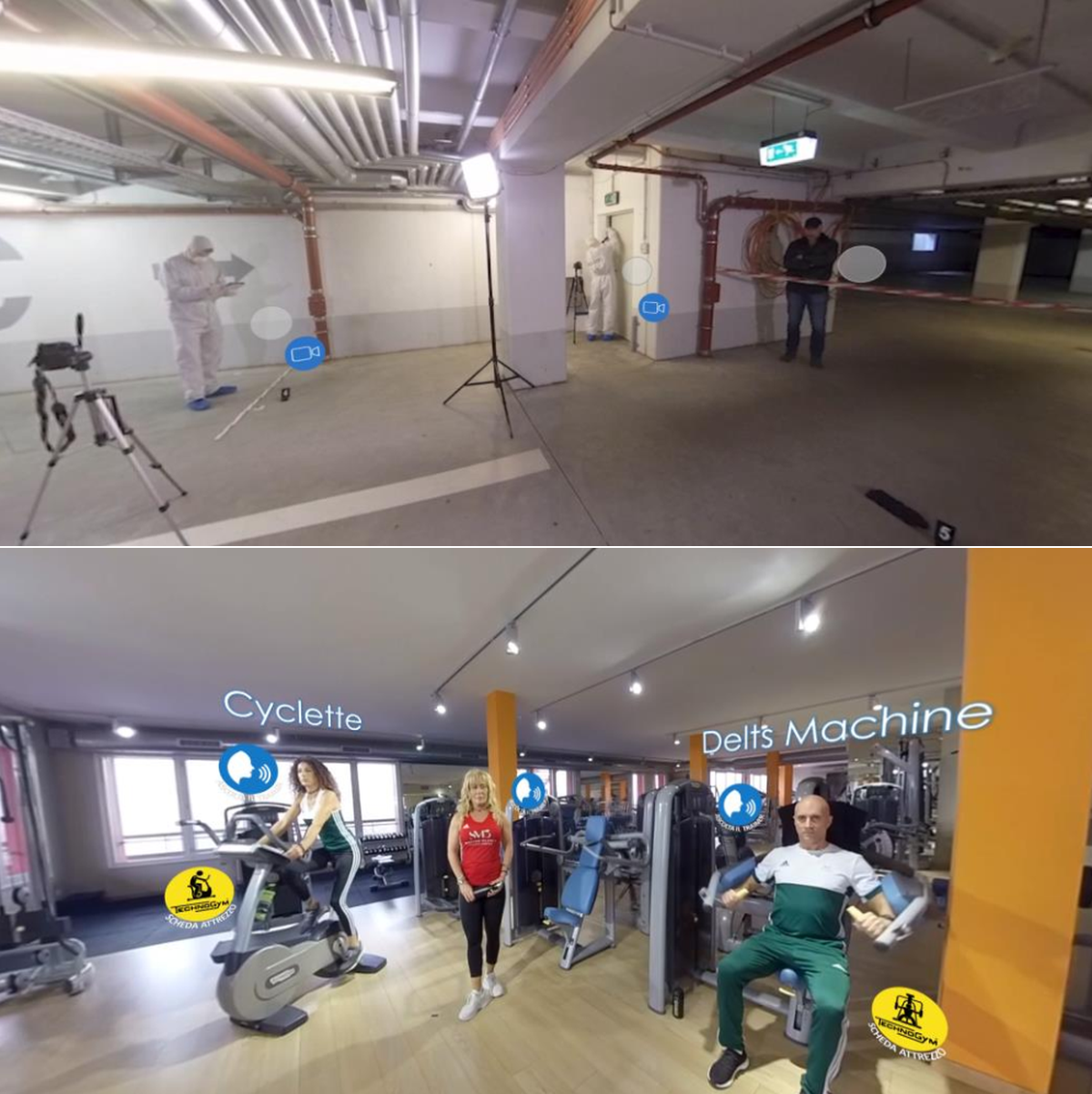}
	\caption{Top shows crime scene pilot by RBB, bottom shows sporting gym pilot by Mediaset. } 
	\label{fig:pilots}
\end{figure}

For both scenarios, first concepts for potential pilots have been developed during joint workshops with the broadcasters' production departments. The most promising concepts have been subsequently refined and transformed into storyboards, which formed the base for the pilot production. Two pilots have been produced by RBB: \emph{Crime Scene} and \emph{Fontane360}. The first one is a CSI-like crime scene investigation with gamification elements, whereas the second one provides a more serene experience and follows Fontane through one of his walks through Brandenburg. RTI produced two pilots named \emph{Sporting Gym} and \emph{Universita Cattolica of Milan}. The former explains the proper use of gym equipment, whereas the latter has the aim of helping new students in choosing the right course.
Figure \ref{fig:pilots} shows the RBB and RTI pilots.

The pilots were used subsequently to assess the immersive experience as a whole, as well as its major components (like the players, 3D mentor, personalisation, automatic camera path). In the following, we will report briefly the setup and outcome of the assessment of the \emph{first} prototypes of the tools (available since Q2 2019). The assessment of the \emph{final} prototypes (which have taken into account the recommendations and criticism mentioned in the assessment of the first prototypes) is still ongoing and therefore cannot be reported yet.

Some of these tests focused on the Hyper360 toolkit, specifically on the Annotation Tool OmniConnect (Web) and the 3D Mentor production suite CapTion. Other tests were made to assess the results of these tools, such as interactive video applications (Web and HbbTV), the Automatic Camera Path, the 3D Mentor and the Personalisation features. All tools and their outputs were tested by RTI and RBB. Tools were evaluated by professionals from the relevant production and innovation teams, the tools outputs were tested by external consumers who had been invited for dedicated assessment sessions.

As the Audience Assessment efforts focused on user acceptance and market relevance, the research questions and methodology were mainly qualitative. The pilot partners RBB and RTI had agreed on the methods and questions with the responsible technical partners in advance, so that the results are easy to compare and summarise. Basically, after a brief introduction test, users were asked to use the services, comment audibly on their expectations and reactions (Thinking-Aloud Method), and finally fill a questionnaire. These questionnaires mainly focused on the consumers’ satisfaction with the provided services, including topics like usability, performance and User Experience.

The feedback comments from the test users concerning the use of the Annotation Tool \textbf{OmniConnect} were very helpful as they helped improving the User Interface as well as the User Experience. There were numerous inputs that have been considered since then, and realised where possible. Especially the design of hotspots needed to be improved, in order to create engaging and appealing applications, as users had considerable difficulties in understanding what they were supposed to expect from interactions with individual buttons. One major criticism by several users was that the interactive application could not be viewed and used in head-mounted displays or at least on smartphones which would allow the user to look around, rather than move the scene across the screen with a mouse. Here the users confirmed an expectation that smartphones and head-mounted displays would make the content applications much more engaging. One interesting aspect in the comparison between the (national) test groups was, that all testers in the RBB assessment criticised that, contrary to what they knew from other players, users have to point the mouse in the direction where they want to move, instead of grabbing the sphere or the background image/video and moving this around, while only a few mentioned the same in the RTI assessment.

For the \textbf{HbbTV player},  the overall impression was that the navigation of the HbbTV app was clear. However, it did not always go smooth, or the application did not always react the way it had been expected. A major point of criticism was focused on the performance. Videos did not run smoothly, but sometimes had a “stop-and-go” feeling and also the fact that additional content of a new scene was already visible when the current scene was still on the screen, caused some criticism.

The \textbf{automatic cinematography} which extracts 2D videos from 360$^\circ$ videos was assessed with three different videos, one at RBB and two at RTI. It was very interesting to see that the results on three different videos (in two different test groups) were significantly different. In summary, it can be said that while the tool indeed has the potential to satisfy some users, there seems to be a lot of room for improvement in the eyes of others.

For the \textbf{3D Mentor} test, consumers were shown 360$^\circ$ videos where 3D Mentor impersonations had been embedded. These had been produced with the CapTion toolset. Overall, the consumers were not overly impressed by the 3D Mentor. Especially the movements seemed unnatural, whereas the appearance of the textures was explicitly praised by some of the testers as more natural than the usual 3D CGI impersonations that they knew.

For the test of the \textbf{personalisation and recommendation}, users were first asked to watch certain videos, so that the system could learn the user’s preferences. In a second round they were asked to visit the same videos again to see, if the system’s recommendation would actually meet their preferences. The results were positive, as by taking into account the recommendations a significantly higher engagement factor of the consumer could be achieved.

The \textbf{overall feedback} from consumers was positive with respect to the concept of interactive 360$^\circ$ video experiences, albeit with some criticism on individual aspects or features. Especially the performance of both the browser and the HbbTV applications were criticised, but the feedback on UI and UX design was very helpful in optimising the way the content partners will create such experiences in the future. It was interesting to see how different the results were, not only comparing respondents in different countries, but even comparing the results of different applications within RTI.  All in all, both the browser and HbbTV applications were rated very positively, but test users also stated explicitly that applications would be even more attractive on smartphones or head-mounted displays (HMD).

\label{sec:pilots}

\section{Conclusion}

The work done so far in the Hyper360 project on tools for capturing, production, enhancement, delivery and consumption of enriched 360$^\circ$ video content has been presented. Furthermore, the first pilots which have produced with these tools and key results of the assessment of these tools have been described.


%

\appendices

\ifCLASSOPTIONcompsoc
  \section*{Acknowledgments}
\else
  \section*{Acknowledgment}
\fi

This work has received funding from the European Union's Horizon 2020 research and innovation programme, grant n$^\circ$ 761934, Hyper360 (``Enriching 360 media with 3D storytelling and personalisation elements''). Thanks to Rundfunk Berlin-Brandenburg and Mediaset for providing the 360\textbf{$^\circ$}~video content.

\ifCLASSOPTIONcaptionsoff
  \newpage
\fi



%
\begin{IEEEbiography}[{\includegraphics[width=1in,height=1.25in,clip,keepaspectratio]{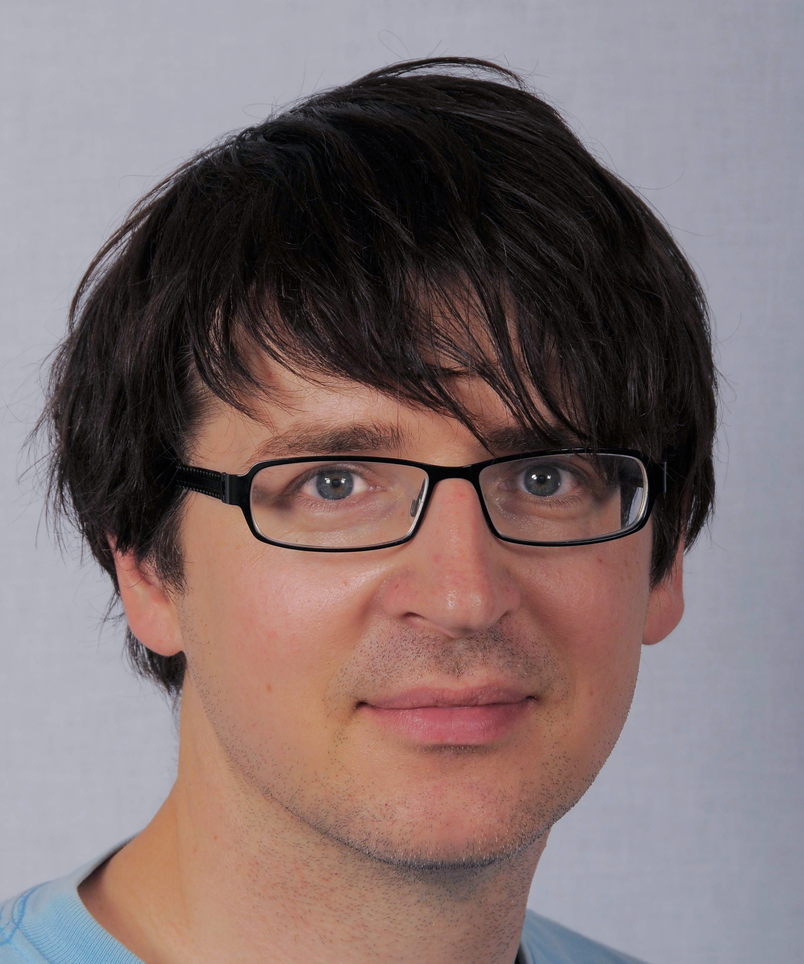}}]{Hannes Fassold}
received a MSc degree in Applied Mathematics  from Graz University of Technology in 2004. Since then he works as a senior researcher at JOANNEUM RESEARCH - DIGITAL. His main research interests are algorithms for digital film restoration and video quality analysis as well as the extraction of semantic metadata (objects, actions, …) from video. 
\end{IEEEbiography}

\begin{IEEEbiography}[{\includegraphics[width=1in,height=1.25in,clip,keepaspectratio]{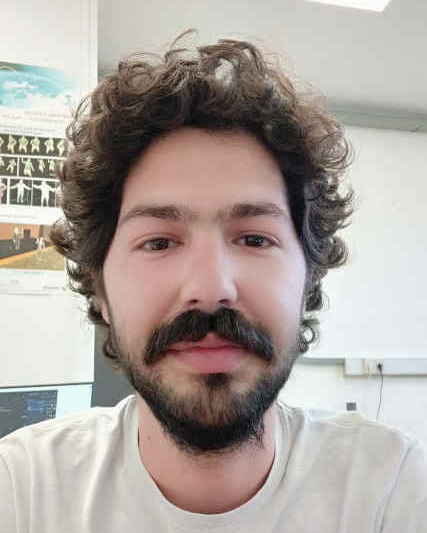}}]{Antonis Karakottas}
obtained his diploma from the Department of Electrical and Computer Engineering, Aristotle University of Thessaloniki in 2015. He has been a member of Information Technologies Institute's (ITI) Visual Computing Lab since July 2017. His main interests are in computer vision, 360 perception, and global illumination, physical based rendering and interactive ray tracing.
\end{IEEEbiography}

\begin{IEEEbiography}[{\includegraphics[width=1in,height=1.25in,clip,keepaspectratio]{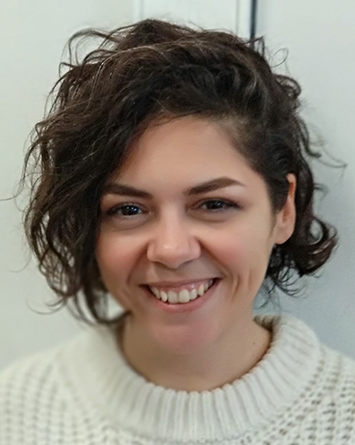}}]{Dorothea~Tsatsou}
is a research associate at the Information Technologies Institute (ITI) of the Centre for Research and Technology Hellas (CERTH) since 2007. Her main research interests include knowledge extraction and representation, automated reasoning, knowledge/ontology engineering, personalization and recommendation. Her involvement with those research areas has led to authoring several scientific papers for international conferences and journals.
\end{IEEEbiography}

\begin{IEEEbiography}[{\includegraphics[width=1in,height=1.25in,clip,keepaspectratio]{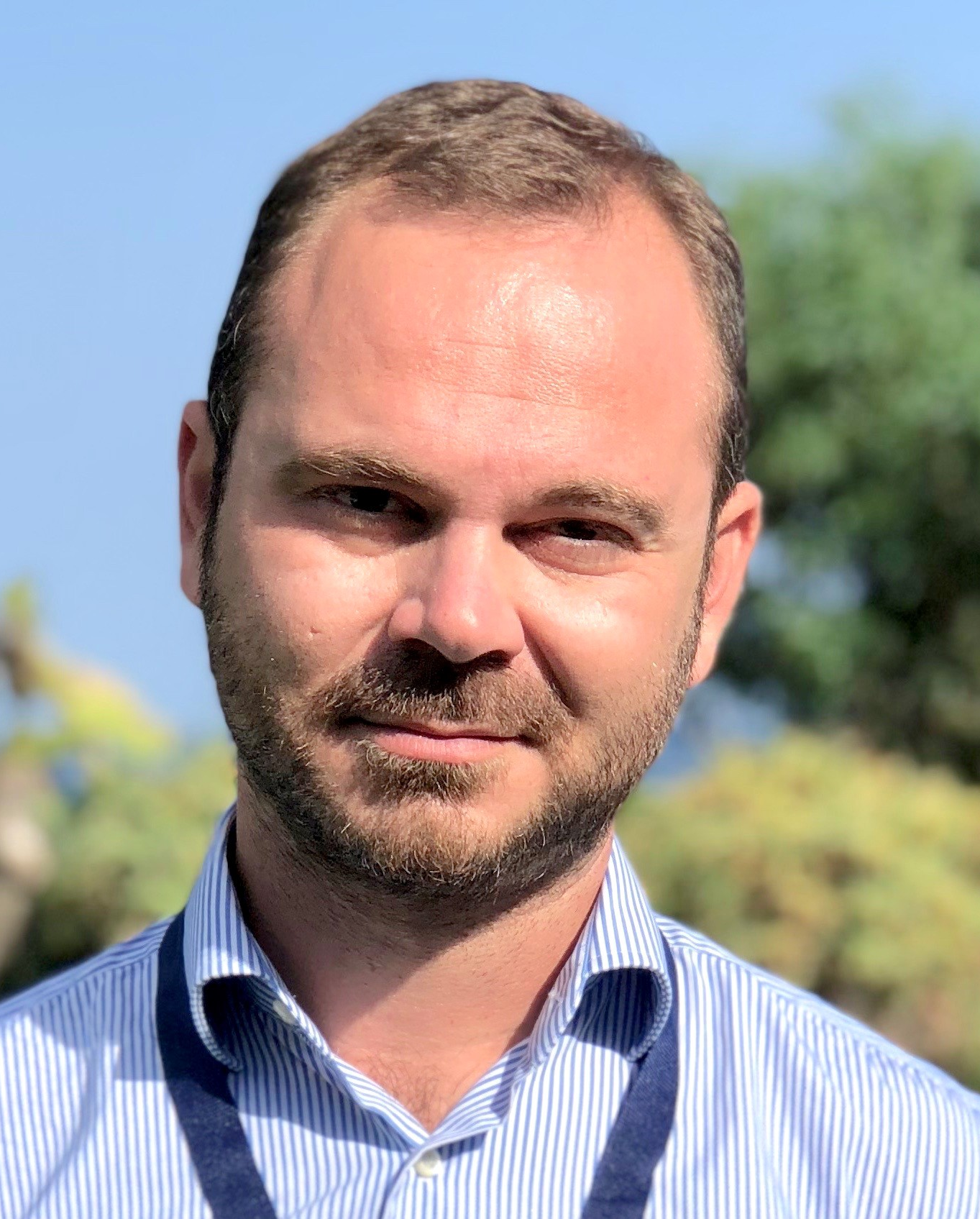}}]{Dimitrios~Zarpalas}
is a Senior Researcher (grade C) at ITI of CERTH. He holds an ECE diploma from Aristotle University of Thessaloniki, an MSc from The Pennsylvania State University, and a PhD from School of Medicine, A.U.Th. He works on 3D/4D computer vision and machine learning: volumetric video, 4D reconstruction of moving humans; 3D medical image processing.
\end{IEEEbiography}

\begin{IEEEbiography}[{\includegraphics[width=1in,height=1.25in,clip,keepaspectratio]{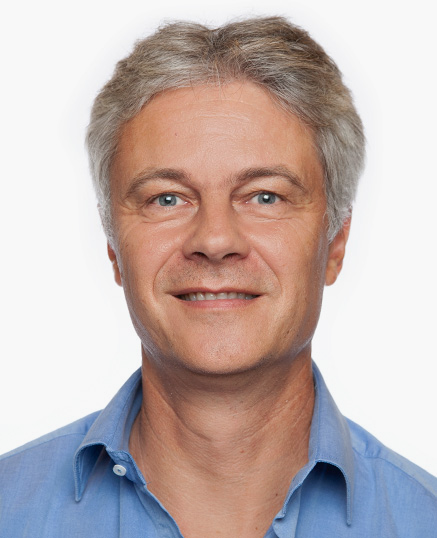}}]{Barnabas Takacs}
is an expert computer scientist in the fields of artificial intelligence, virtual- and augmented reality (XR), human modeling and animation, computer vision, tracking, real-time image processing, face recognition, security systems and novel medical technologies. He worked on virtual humans, MMW radars, medical simulation, VR 360 experiences, AR Systems for Industry and Entertainment and also Hollywood film productions. 
\end{IEEEbiography}

\begin{IEEEbiography}[{\includegraphics[width=1in,height=1.25in,clip,keepaspectratio]{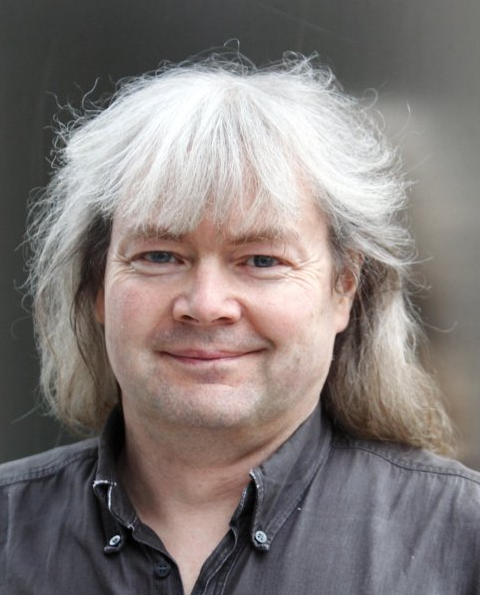}}]{Christian Fuhrhop}
received his diploma in computer science from the Technical University Berlin in 1987. At Fraunhofer he has participated in several EU-funded projects, including webinos, GlobalITV and Producer and has coordinated the MPAT project. He participated in TV Anytime standardization and was active in industry projects related to TV based applications. 
\end{IEEEbiography}

\begin{IEEEbiography}[{\includegraphics[width=1in,height=1.25in,clip,keepaspectratio]{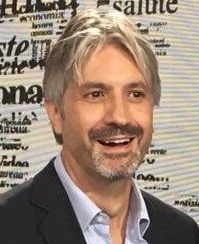}}]{Angelo Manfredi}
has worked for more than ten years in the  TELCO sector, as a Projects Manager for Italian and foreign Customers. From 2017 he is working as Project coordinator in the European “Hyper360” project regarding the enrichment of 360$^\circ$ videos. 
\end{IEEEbiography}

\begin{IEEEbiography}[{\includegraphics[width=1in,height=1.25in,clip,keepaspectratio]{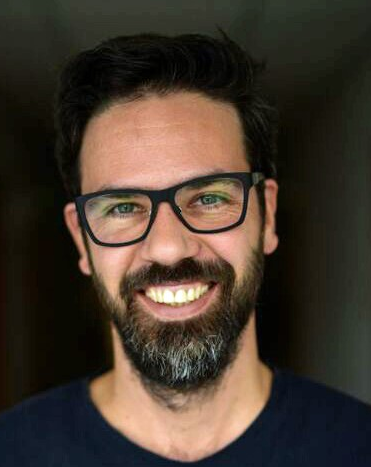}}]{Nicolas Patz}
was awarded his M.A. in English and History of Art from Freie Universität Berlin. He has more than 16 years of experience in EC funded research projects, specialising in user-friendly interaction concepts. At RBB Nico has focused on concepts based on access and interactivity related to broadcast and hybrid services. 
\end{IEEEbiography}

\begin{IEEEbiography}[{\includegraphics[width=1in,height=1.25in,clip,keepaspectratio]{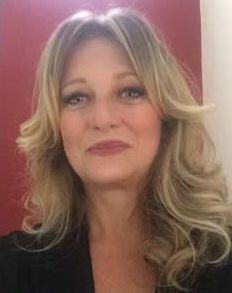}}]{Simona Tonoli}
has a multi-year experience in the Italian TV sector as Business Development, CRM and Digital Marketing, Digital transformation and  Innovation projects on Pay DTT TV and OTT TV.  She is currently in charge of European Project Innovation at national Italian TV Mediaset RTI including  360 Virtual Reality, Augmented Reality and HBBtv projects.
\end{IEEEbiography}

\begin{IEEEbiography}[{\includegraphics[width=1in,height=1.25in,clip,keepaspectratio]{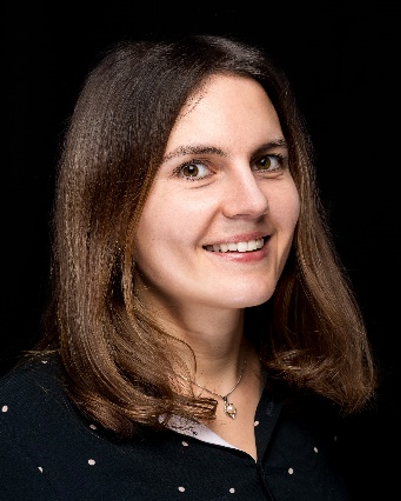}}]{Iana Dulskaia}
is a Senior Consultant at Eurokleis srl with an experience in developing EU funded projects in fields of ICT, smart mobility and Future and Emerging Technologies. She received her Ph.D. in Management, Banking and Commodity Sciences from La Sapienza university of Rome. Her fields of expertise are in business development, technological transfer and innovation management.
\end{IEEEbiography}

\vfill






\bibliography{IEEEfull}
\bibliographystyle{IEEEtran}

\end{document}